# Tea Chrysanthemum Detection under Unstructured Environments Using the TC-YOLO Model

Chao Qi [a], Junfeng Gao [b, &], Simon Pearson [b], Helen Harman [b], Kunjie Chen [a, *], Lei Shu [a, *]

[a] *College of Engineering, Nanjing Agricultural University, Nanjing 210031, China.*

[b] *Lincoln Agri-Robotics Centre, Lincoln Institute for Agri-food Technology, University of Lincoln, Lincoln, Riseholme Park, LN2 2LG, UK.*

[&]*These authors contributed equally to this work and should be considered co-first authors.*

* Corresponding author.
*E-mail addresses:* chaoqi.njau@gmail.com (Chao Qi), jugao@lincoln.ac.uk (Junfeng Gao), spearson@lincoln.ac.uk (Simon Pearson), hharman@lincoln.ac.uk (Helen Harman), kunjiechen@njau.edu.cn (Kunjie Chen), lei.shu@njau.edu.cn (Lei Shu).

**Abstract:** Tea chrysanthemum detection at its flowering stage is one of the key components for selective chrysanthemum harvesting robot development. However, it is a challenge to detect flowering chrysanthemums under unstructured field environments given the variations on illumination, occlusion and object scale. In this context, we propose a highly fused and lightweight deep learning architecture based on YOLO for tea chrysanthemum detection (TC-YOLO). First, in the backbone component and neck component, the method uses the Cross-Stage Partially Dense Network (CSPDenseNet) as the main network, and embeds custom feature fusion modules to guide the gradient flow. In the final head component, the method combines the recursive feature pyramid (RFP) multiscale fusion reflow structure and the Atrous Spatial Pyramid Pool (ASPP) module with cavity convolution to achieve the detection task. The resulting model was tested on 300 field images, showing that under the NVIDIA Tesla P100 GPU environment, if the inference speed is 47.23 FPS for each image (416 × 416), TC-YOLO can achieve the average precision (AP) of 92.49% on our own tea chrysanthemum dataset. In addition, this method (13.6M) can be deployed on a single mobile GPU, and it could be further developed as a perception system for a selective chrysanthemum harvesting robot in the future.

# 1. Introduction

Current studies show that tea chrysanthemums have significant commercial value (Liu et al., 2020; Liu et al., 2019). Not only that, but tea chrysanthemums can offer a range of health benefits (Hou et al., 2017; Yue et al., 2018). For example, they can significantly inhibit the activity of carcinogens and have distinct anti-aging, cholagogic and antihypertensive effects (Zheng et al.,

2021). In addition, apigenin and naringenin in tea chrysanthemums can effectively reduce or even reverse the prevalence of obesity (Thaiss et al., 2016). In nature environments, a tea chrysanthemum plant could present multiple flower heads varying in different stages and sizes. The examples of different flowering stages of chrysanthemums are shown in Fig. 1.

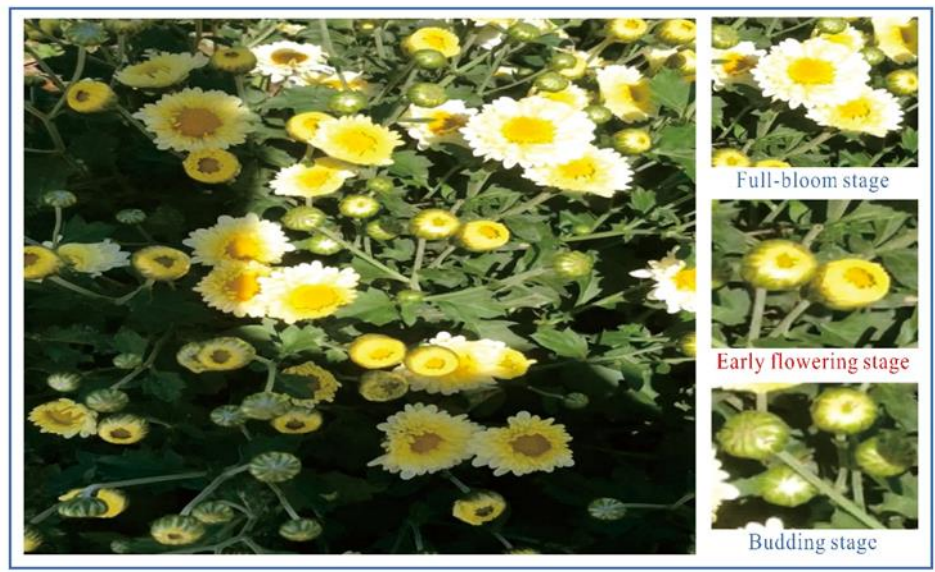

**Fig. 1.** A typical tea chrysanthemum plant with three different flowering stages.

At present, tea chrysanthemum is mainly manually harvested at the early flowering stage, and this process is labor-intensive and time-consuming. The development of artificial intelligence and selective harvesting robots will help to prevent crop wastage due to the current shortages in skilled labourers (Li et al., 2021). Therefore, it is highly needed to develop an automated selective harvesting robot. Typically, a robotic harvesting process is divided into two steps. First, a computer vision system is used to detect the objects of interest. Then the manipulator moves towards the detected objects and the gripper harvests the objects guided by the detection results. Accurately detecting the chrysanthemum at the early flowering stage is critical for guiding the manipulator to the correct position. Although methods based on machine learning technology have made remarkable achievements in agricultural applications (Gao et al., 2020), it is still difficult to develop a lightweight network for a selective harvesting robot under unstructured environments.

The study on chrysanthemum detection based on machine vision began as early as 1996. Kondo et al. developed a system for cutting chrysanthemums. The chrysanthemums were detected using features the direction, position, size, and shape for cutting chrysanthemums. Unfortunately, this study failed to provide definite detection results (Kondo et al., 1996). Subsequently, Warren developed a system for identifying chrysanthemum leaves through extracting image features, including the length and width of the leaves, and the shape of the leaf apices. For the system, the precision of detection was up to 75%, and generally, its running speed was roughly the same as the manual running speed (Warren, 2000). Tarry et al. developed an integrated system for chrysanthemum bud detection by extracting color characteristics. The system can be utilized to automate labor-intensive work in flower greenhouses with a detection precision of 78.2% (Tarry et al., 2014). Tete et al. extracted the characteristics of chrysanthemum leaves using two algorithms, i.e., threshold segmentation and K-means clustering, and detected the leaves infected with pests and diseases. However, this study merely provided the images of segmented chrysanthemums without definite qualitative results (Tete&Kamlu, 2017) . Yuan et al. extracted the characteristics of chrysanthemum petals through the random forest algorithm and tested the chrysanthemums for variety. The experimental results showed that the detection precision increased with the number of extracted characters, and AP was up to 95% at maximum (Yuan et al., 2018). Yang et al. segmented the White Chrysanthemums based on HSV color space, extracted target image features from an irrelevant background, and developed a robot plucking system for Hangzhou White Chrysanthemum, of which the precision for identification reached 85%, and the average plucking time was 0.4s. However, the robustness for the extraction method based on color features is poor

(Yang et al., 2018). Considering the shortcomings, Yang et al. further extracted the color and texture features of the images by the RGB value and the gray level co-occurrence matrix. The experimental results have shown that the system can effectively detect the image of Hangzhou White Chrysanthemum under different light conditions with average plucking time of 0.7s and plucking success rate of 90%. This study added the chrysanthemum texture features based on the original study. The detection precision can be effectively improved by the extraction and reasonable utilization of image features, but the speed for inference is still to be further improved (Yang et al., 2019).

There are advantages in the detection precision for the chrysanthemum detection method based on traditional machine learning. However, in a complicated unstructured environment, it is difficult to detect the chrysanthemum images, and the actual speed for detection is too slow to meet the requirements for large-scale operations. Deep learning, a subset of machine learning, enables learning of hierarchical representations and the discovery of potentially complex patterns from large data sets. It has shown impressive advancements on various problems in natural language processing and computer vision, and the performance of deep convolutional neural networks (CNNs) on image classification, segmentation and detection are of particular note (Togacar et al., 2020). Deep learning in the agriculture domain is also a promising technique with growing popularity (Gao et al., 2021). Liu et al. proposed a deep learning model based on VGG16 and ResNet50 networks for identifying big chrysanthemums. Relevant experimental results have shown that the precision of the model may be up to 78%, and the inference speed for detecting a chrysanthemum image can reach 10ms (Liu et al., 2019). Liu tested the flowering periods of 7 commercial chrysanthemum species with the DNN algorithm, and the detection precision of the model reached 96% (Liu et al., 2020). Van detected the chrysanthemum image with noise with Faster R-CNN algorithm, by which the detection precision of the model was up to 76%, and the inference speed for detecting a chrysanthemum image was 0.3s (Van Nam, 2020). The above studies have shown that deep learning approaches outperform the traditional machine learning methods regarding to inference speed. However, for these studies, the chrysanthemum detection was carried out under controlled environments, and there is still a gap to reach a reliable detection in the real world. A CNN architecture was designed to understand common features of these environments, and to take advantage of the synergy between them. In many works, fusing features of different scales is an important means of improving detection performance and computational speed. Low-level features have a high resolution and contain more details, but due to few convolution operations, these features have less semantic information and more noise. In contrast, high-level features have more semantic information, but have a low resolution and poor ability to perceive details (Liu et al., 2021). Apparently, the key to improving the performance of detection models is to efficiently fuse the features of different convolutional layers.

With the rapid development of deep learning technology, the superiority of feature fusion algorithms has become increasingly prominent. Feature fusion algorithms can be divided into three categories: algorithms based on Bayesian decision theory, algorithms based on sparse representation theory, and algorithms based on deep learning theory. In object detection tasks, the main algorithms employed are based on deep learning theory, i.e. the fusion of multi-class features from different neural networks, such as spatial pyramid pooling (SPP) structures and atrous spatial pyramid pooling (ASPP). Feature fusion algorithms improve detection accuracy through obtaining numerous features of small objects (Bae et al., 2020). However, extending the architecture usually requires more computation. This decreases the model inference speed, resulting in low detection efficiency.

The task of harvesting chrysanthemum at a specific maturity stage usually requires shortening the reasoning time on small devices, which poses a complex challenge for computer vision algorithms. In response to this, a lightweight network based on feature fusion is proposed in this paper, which has a potential to be deployed on the embedded GPU chip without compromising performance. The primary aim for the network design is to achieve more diverse combinations of gradients as well as to reduce the computational workload. The contributions of this paper are as follows:

1. A lightweight object detection model was designed, using CSPDenseNet as the main network and applying a few feature fusion modules. This model can achieve real-time detection work.

2. The impact of dataset sizes, different data enhancement methods, and complex unstructured scenarios were quantified with the TC-YOLO model, and the superiority of the TC-YOLO model was validated by comparing it with several state-of-the-art object detection models.

3. A lightweight detection architecture was designed to detect chrysanthemums at the early flowering stage, that can adapt to complex unstructured environments (light changes, shading and overlap).

The organization of this paper is as follows. Section 2 describes the materials and methods, especially the settings of the dataset and the design details of the proposed TC-YOLO network. Section 3 introduces the experimental results of TC-YOLO. Section 5 discusses the contribution of this paper, the limitations of the study and unresolved problems. We reflect on these problems and indicate the future possible solutions. Finally, a brief summary and conclusions of this paper are provided in Section 6.

## 2. Materials and methods

*2.1. Dataset*

The chrysanthemum datasets used in this paper were collected from chrysanthemum breeding bases in Sheyang County in Jiangsu Province, Dongzhi County in Anhui Province, and Nanjing Agricultural University in Jiangsu Province from October 2019 to October 2020. An Apple X phone camera was used to capture images with a resolution of 1080 × 1920. All images were taken under natural light, including variations in background, illumination, growth stages, occlusion, and overlapping.

Chrysanthemums were captured in these images at three stages: the budding stage, the early flowering stage, and the full-bloom stage. The budding stage refers to the stage when buds on the main stem or branch of the chrysanthemum plant can be identified by the naked eye and the petals are not opened. The early flowering stage refers to the stage when the petals are not fully opened, and the full bloom stage refers to the stage when the petals are fully opened. A total of 1000 chrysanthemum image samples were divided into training, validation, and test datasets at the ratio of 6:3:1. Each prediction module had three prior anchors of different scales. Through *k*-means clustering [39], nine prior anchors — (10,13), (16,30), (33,23), (30,61), (62,45), (59,119), (116,90), (156,198), and (373,326) — of the chrysanthemum dataset were obtained. The pre-processed chrysanthemum image 416 × 416 × 3 was transformed into a 208 × 208 × 12 feature map via the focus slicing operation, and then one convolution operation of 32 convolution kernels was performed to finally form a 208 × 208 × 32 feature map. The preprocessed chrysanthemum images are shown in Fig. 2.

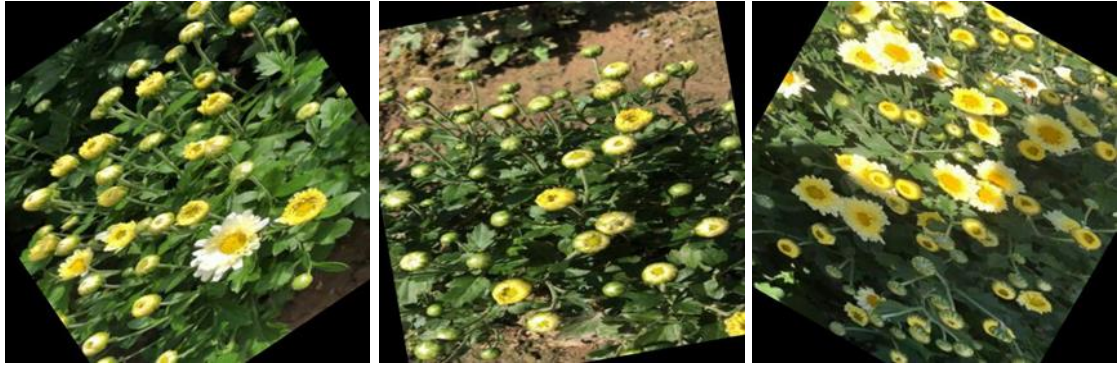
**Fig. 2.** Preprocessed chrysanthemum images. The preprocessed methods include flip and rotation.

## 2.2. TC-YOLO

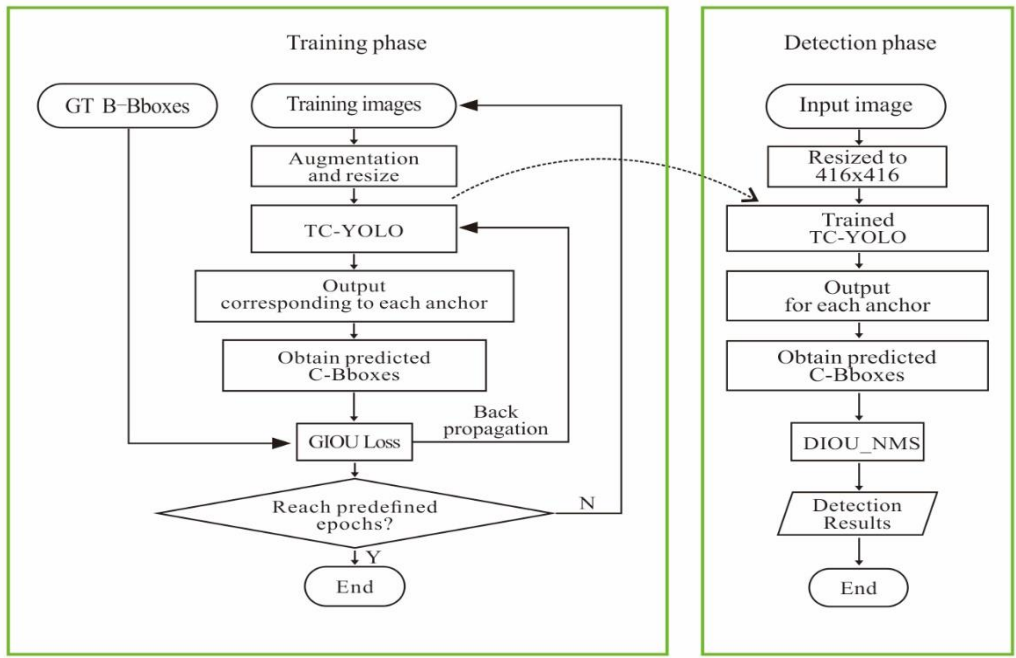

**Fig. 3.** The flow chart of TC-YOLO training and detection. In the training phase, the original chrysanthemum image is enhanced and the image size is adjusted to 416*416. The coordinates of the predicted boxes are obtained after the image passes through the proposed TC-YOLO framework. The GIOU Loss function and back-propagation method are used to train the network. The training stops when the preset epoch is reached. In the detection phase, the DIOU_NMS is used to evaluate the distance between the predicted box and the ground true box, so as to output accurate detection results.

TC-YOLO is a lightweight CNN (13.6M) that can adapt to complex unstructured environments (illumination variation, occlusion, overlap, etc.). Fig. 3 shows the flow chart of TC-YOLO training and detection. TC-YOLO has three main components: the backbone, the neck and the head component. In the backbone component and the neck component, we we used CSPDenseNet as the main network, applying both the CBL module and the SPP module. In the Head component, the multi-scale fusion network is the Recursive Feature Pyramid (RFP)+ Atrous Spatial Pyramid Pooling (ASPP) structure. Furthermore, the Mish activation function and the GIOU loss function were used throughout the network. The network structure is shown in Fig. 4.

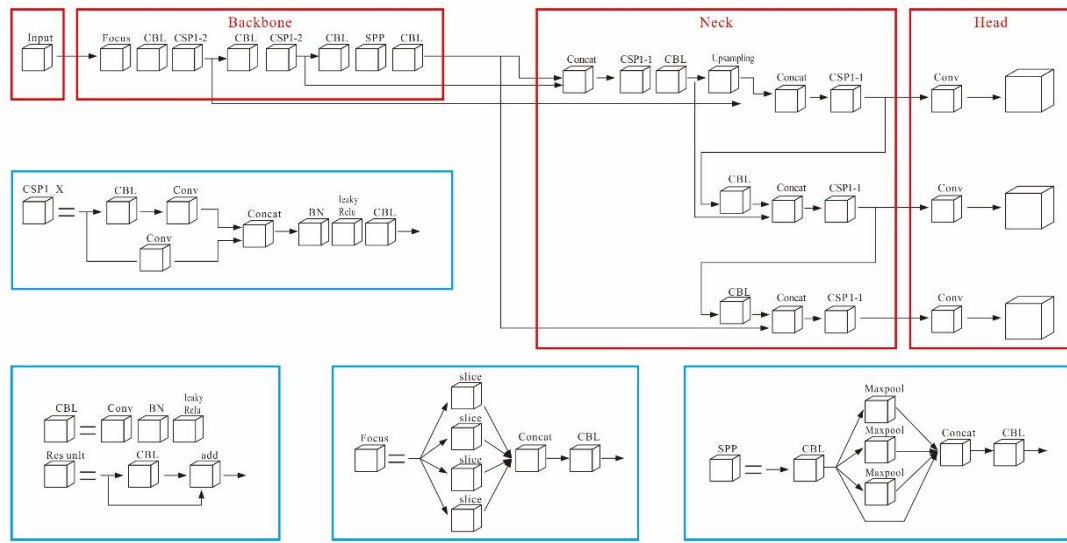

**Fig. 4.** Structure of the proposed TC-YOLO network. Convolution + Batch normalization + Leaky ReLU (CBL), Cross-Stage Partial (CSP), Spatial Pyramid Pooling (SPP), Batch Normalization (BN).

*2.2.1. Backbone*

*2.2.1.1. CSPDenseNet*

The CSPDenseNet module employed in this section retains the advantages of the DenseNet feature reuse feature, while preventing excessive repeat gradient information by truncating the gradient flow. This module consists of the partially dense block and the partially transition layer. In the partially dense block, the number of gradient paths can be doubled by splitting and merging operations, which can alleviate the drawback of using explicit feature map replication for splicing due to the cross-stage strategy. Moreover, as the number of channels involved in the underlying dense layer operations in some partial dense blocks is only half of the original number, nearly half of the computational bottleneck can be effectively eliminated. The CIO of the dense block is calculated as follows:

$$(c*m) + \frac{(m^2+m)*d}{2} \qquad (1)$$

where $c$ represents the number of channels in the basic feature map of the dense block, $m$ stands for the number of dense layers and $d$ refers to the growth rate. Although m and d are typically much smaller than c, a partially dense module can still save up to half of the memory flow in the network.

The CIO of partial dense blocks is calculated as follows:

$$\frac{(c*m)+(m^2+m)*d}{2} \qquad (2)$$

In the partial transition layer, after the output of the dense layer passes through one transition layer, the output $x_T$ will be concatenated with $x_0''$ and then pass through another transition layer before generating the output $x_U$. The feedforward pass and weight update of the module are shown in Eq. (1) and Eq. (2), respectively.

$$x_k = w_k * [x_0, x_1, ..., x_{k-1}] \qquad (3)$$

$$x_T = w_T * [x_0, x_1, ..., x_k]$$

$$x_U = w_U * [x_0, x_T]$$

$$w_k' = f(w_k, g_0, g_1, g_2, ..., g_{k-1}) \qquad (4)$$

$$w_T' = f(w_T, g_0, g_1, g_2, ..., g_k)$$

$$w_U' = f(w_U, g_0, g_T)$$

where * is the convolution operator; $[x_0, x_1, ...]$ is the connection $x_0, x_1, ..., w_i$; $x_i$ stands for the weight and output of the $i$-th dense layer; $f$ is the function of weight updating; and $g_i$ is the gradient propagating to the $i$-th dense layer.

*2.2.1.2. CBL + SPP*

The CBL comprises a convolutional layer, batch normalisation (BN) and a Leaky ReLU activation function. To maximise the difference in gradient flow through the CBL, an SPP component is embedded at the output of the CBL component for the tandem operation. To be specific, the equation for the CBL+SPP component is as follows:

$$K_h = ceil\left(\frac{h_{in}}{n}\right) \tag{5}$$

$$S_h = ceil\left(\frac{h_{in}}{n}\right)$$

$$P_h = floor\left(\frac{k_h*n - h_{in} + 1}{2}\right)$$

$$H = 2*P_h + h_{in}$$

$$K_w = ceil\left(\frac{w_{in}}{n}\right) \tag{6}$$

$$S_w = ceil\left(\frac{w_{in}}{n}\right)$$

$$P_w = floor\left(\frac{k_w*n - w_{in} + 1}{2}\right)$$

$$w = 2*P_w + w$$

where $K_h$, $h$, $P_h$, and $S_h$ represent the height of the kernel, the height of the feature mapping matrix, the number of fillings in the height direction of the feature mapping matrix, and the step size in the height direction of the feature mapping matrix, respectively; *ceil* () stands for the rounding up symbol; *floor* () denotes the rounding down symbol; and $h_{in}$ refers to the height of the input data.

According to Eq. (5) and Eq. (6), the equation of the feature mapping matrix is as follows:

$$\left[\frac{h + 2p - f}{s} + 1\right] * \left[\frac{w + 2p - f}{s} + 1\right] \tag{7}$$

where *p* represents padding, *s* denotes stride, and *f* indicates the input data size.

*2.2.2. Head*

The Head component is the prediction part of the network, and the scale of the final predicted feature map is 76 × 76, 38 × 38, and 19 × 19. The structure of the prediction part is Recursive Feature Pyramid (RFP). RFP adds feedback connections to FPN as highlighted in Fig. 5. $R_i$ stands for the feature transformations before connecting them back to the bottom-up backbone. Then, $\forall i = 1,…,S,$ the output feature $f_i$ of RFP is defined by

$$f_i = F_i(f_{i+1}, X_i), X_i = B_i(X_{i-1}, R_i(f_i)) \qquad (8)$$

Which makes RFP a recursive operation. We unroll it to a sequential network, *i.e.*, $\forall i = 1, ..., S, t = 1, ...T,$

$$f_i^t = F_i^t(f_{i+1}^t, X_i^t), X_i^t = B_i^t(X_{i+1}^t, R_i^t(f_i^{t-1})) \qquad (9)$$

where $T$ is the number of unrolled iterations, and we use superscript $t$ to denote operations and features at the unrolledstep $t$. $f_i^0$ is set to 0. In our implementation, $F_i^t$ and $R_i^t$ are shared across different steps.

We utilised Atrous Spatial Pyramid Pooling (ASPP) to achieve the connecting module R, that takes a feature $f_i^t$ as its input and transforms it to the RFP feature used in Fig. 5. In this module, there are four parallel branches that take $f_i^t$ as their inputs, the outputs of which are then concatenated together along the channel dimension to form the final output of R. Three branches of them employ a convolutional layer followed by a ReLU layer, the number of the output channels is 1/4 the number of the input channels. The last branch uses a global average pooling layer to compress the feature, followed by a 1x1 convolutional layer and a ReLU layer to transform the compressed feature to a 1/4-size (channel-wise) feature. Finally, it is resized and concatenated with the features from the other three branches. The convolutional layers in those three branches are of the following configurations: kernel size = [1, 3, 3], atrous rate =[1, 3, 6], padding = [0, 3, 6]. Unlike the original ASPP, we do not have a convolutional layer following the concatenated features as in here R does not generate the final output used in dense prediction tasks. Note that each of the four branches yields a feature with channels 1/4 that of the input feature, and concatenating them generates a feature that has the same size as the input feature of R.

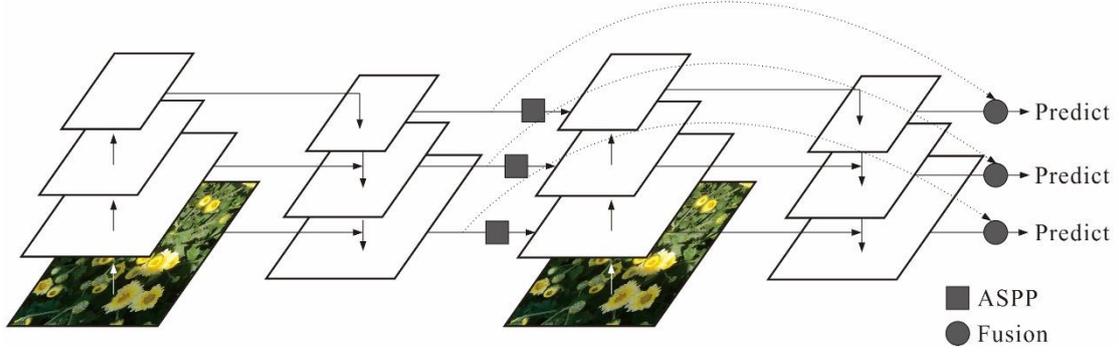

**Fig. 5.** The architecture of Recursive Feature Pyramid (RFP).

*2.3. Evaluation metrics*

For object detection applications, average precision (AP) is a standard metric for evaluation of model performance. In our case, we calculated the average accuracy for one category, tea chrysanthemums, to check the overall performance of the model. Precision is a ratio of true object detections to the total number of objects that a model predicted. Recall is a ratio of true object detections to the total number of objects in the dataset. The equation is as follows:

$$\text{AP} = \sum_{k=1}^{N} P(k) \Delta recall\ (k) \tag{10}$$

where *N* is the total number of images in the test dataset, *P(k)* is the precision value at *k* images and *recall (k)* is the change of the recall between *k* and *k* − 1 images.

*2.4. Experimental Setup*

Four experiments were designed to test the performance of the proposed detection model. In the first experiment (Section 3.1), we randomly selected and established eight datasets in varied sizes to investigate the influence of dataset size for TC-YOLO modeling. In the second experiment (Section 3.2), in order to understand the impact of data augmentation on TC-YOLO, based on the results of the first experiment, we selected 14 data augmentation methods and performed the test using different configuration schemes. In the third experiment (Section 3.3), in order to investigate the robustness of TC-YOLO in complex unstructured scenarios, we set up nine unstructured scenarios based on our previous chrysanthemum harvesting experience. In the fourth experiment (Section 3.4), in order to thoroughly study the detection performance of TC-YOLO, we tested nine latest target detection frameworks on the chrysanthemum dataset.

The experiments were carried out on a server with NVIDIA Tesla P100, CUDA 11.2. The operating system was Ubuntu 16. The basic detection framework was CSPDenseNet. In the training process, the key hyperparameter settings were as follows: learning rate = 0.01; momentum = 0.937; gamma = 0.1; weight decay = $5 \times 10^{-4}$; warmup_momentum = 0.8; warmup_bias_lr = 0.1. The optimizer used was a stochastic gradient descent (SGD).

## 3. Results

### 3.1. Impact of Dataset Size on the TC-YOLO

To verify the impact of the dataset size on the chrysanthemum detection task, eight datasets of varied sizes were randomly selected from the chrysanthemum training set, which contain 50, 100, 200, 500, 1000, 2000, 3000, 4000 images, respectively. The AP of the chrysanthemum is shown in Fig. 6.

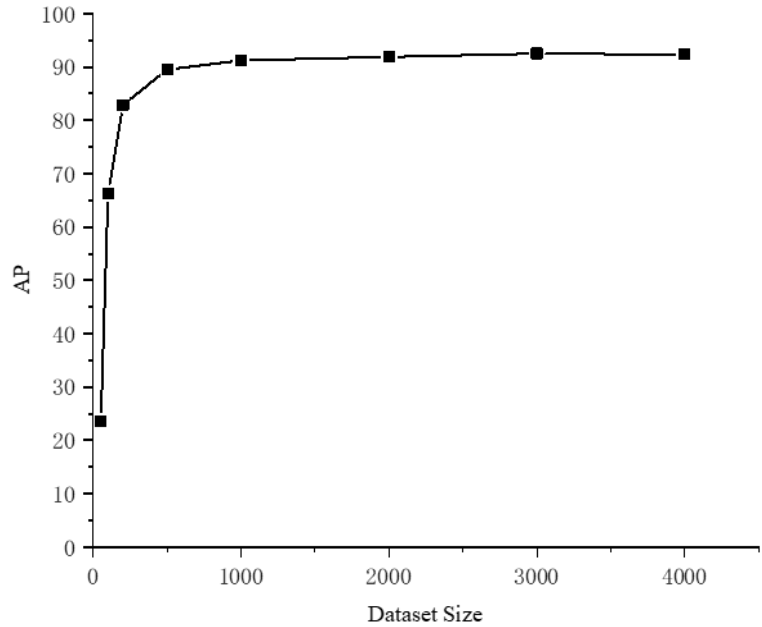

**Fig. 6.** Impact of dataset size on the detection task

It can be seen that the performance of the detection model improves with the increase in dataset size in Fig. 6. When the number of images was less than 500, the AP value increased rapidly (23.62% to 66.29%, improved by 180.65%) with the increase in the number of images. When the number of samples reached at 1000, only small improvements can be obtained (from 91.22% to 92.49%) by adding more samples. The performance of the detection model converged with AP value at 93% after 3000 images. The optimal dataset size was set as 3,000 images for the experiment in the following sections due to its high AP value achieved under the minimal sample number required.

*3.2. The impact of data augmentation on the TC-YOLO*

To explore the influence of data augmentation on TC-YOLO, we selected 14 data augmentation methods for the test, and the test results are shown in Table 1. Our experimental design is to configure these data augmentation methods in turn, and select the best-performing data augmentation methods according to the performance of corresponding models. Surprisingly, the most advanced data enhancement methods, such as Cutout, Mixup, Cutmix and Mosaic, only achieved performance APs of 86.11%, 85.16%, 86.51% and 88.62% respectively. This may be because massive redundant gradient information will significantly reduce the learning ability of the network. We found that Flip and Rotation performed best, with 91.02% and 91.83% AP values, respectively. Moreover, when both Flip and Rotation were configured, the performance of the model was slightly improved (AP=92.49%). Therefore, this combination was adopted for data augmentation. Note that other than Flip and Rotation, the improvement of model performance gained by using other data augmentation methods was rather limited.

**Table 1**

The impact of data augmentation on the TC-YOLO.

| Flip | Shear | Crop | Rotation | Grayscale | Hue | Saturation | Exposure | Blur | Noise | Cutout | Mixup | Cutmix | Mosaic | AP |
|---|---|---|---|---|---|---|---|---|---|---|---|---|---|---|
| √ | | | | | | | | | | | | | | 91.02 |
| | √ | | | | | | | | | | | | | 87.96 |
| | | √ | | | | | | | | | | | | 87.63 |
| | | | √ | | | | | | | | | | | 91.83 |
| | | | | √ | | | | | | | | | | 85.11 |
| | | | | | √ | | | | | | | | | 88.67 |
| | | | | | | √ | | | | | | | | 88.23 |
| | | | | | | | √ | | | | | | | 84.52 |
| | | | | | | | | √ | | | | | | 83.26 |
| | | | | | | | | | √ | | | | | 88.89 |
| | | | | | | | | | | √ | | | | 86.11 |
| | | | | | | | | | | | √ | | | 85.16 |
| | | | | | | | | | | | | √ | | 86.51 |
| | | | | | | | | | | | | | √ | 88.62 |
| √ | | | √ | | | | | | | | | | | 92.49 |

*3.3. Impact of Different Unstructured Scenarios on the TC-YOLO*

This study examined the robustness of the proposed model in different unstructured environments, including strong light, weak light, normal light, high overlap, moderate overlap, normal overlap, high occlusion, moderate occlusion, and normal occlusion. There were 30236 chrysanthemums in the early flowering stage in nine unstructured environments. Note that there is no strict criterion for reference as defining various scenarios. Some criteria are set artificially for the convenience of testing. First of all, most scenarios may be differentiated empirically, as shown in Fig. 7, which has provided nine scenarios. Additionally, according to our definition, tea chrysanthemums are counted separately in different scenarios, e.g., for a tea chrysanthemum in normal light, normal overlap, and normal occlusion scenarios. In counting, the number of tea chrysanthemums is increased by 1 respectively in the three scenarios.

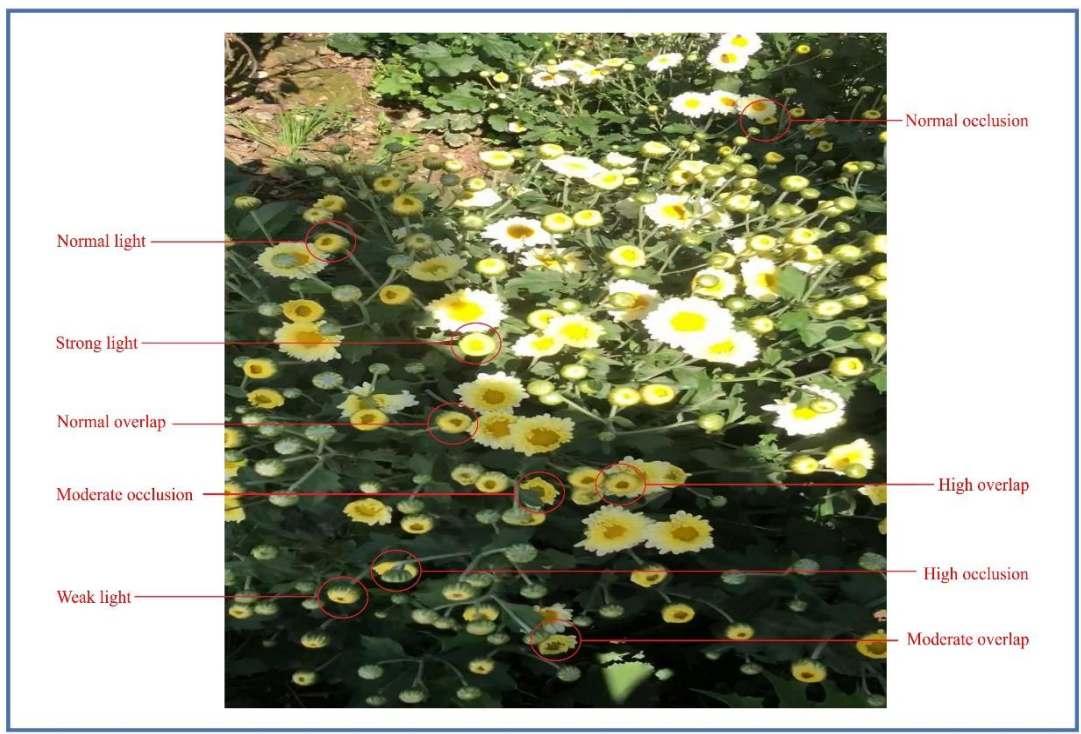

**Fig. 7.** Example of 9 unstructured scenarios.

Table 2 shows that, under normal environmental conditions, the AP of chrysanthemums at the early flowering stage reached impressive values of 96.52%, 97.11%, and 96.88% for normal light, normal overlap and normal occlusion, respectively. When the unstructured environment became complex, the AP of chrysanthemums at the early flowering stage decreased significantly, especially under the scenario of strong light, where the AP was only 80.54%. Interestingly, in all unstructured environments, the error rate under strong light was the highest, reaching 7.54%, while the miss rate under high overlap was the highest, reaching 14.22%. In addition, in general, overlapping had the least influence on the detection of chrysanthemums at the early flowering stage. Under high overlap, the AP, error rate, and miss rate were 82.51%, 3.27%, and 14.22%, respectively. Illumination had the biggest impact on the detection of chrysanthemums at the early flowering stage. With strong light, the accuracy, error rate and miss rate were 80.54%, 7.54%, and 11.92%, respectively.

**Table 2**

Impact of Different Unstructured Scenarios on the TC-YOLO.

| Environment | Count | Correctly Identified | | Falsely Identified | | Missed | |
| --- | --- | --- | --- | --- | --- | --- | --- |
| | | Amount | Rate (%) | Amount | Rate (%) | Amount | Rate (%) |
| Strong light | 6853 | 5519 | 80.54 | 517 | 7.54 | 817 | 11.92 |
| Weak light | 16511 | 14639 | 88.66 | 1174 | 7.11 | 698 | 4.23 |
| Normal light | 23542 | 22723 | 96.52 | 438 | 1.86 | 381 | 1.62 |
| High overlap | 5699 | 4702 | 82.51 | 186 | 3.27 | 811 | 14.22 |
| Moderate overlap | 12998 | 11814 | 90.89 | 476 | 3.66 | 708 | 5.56 |
| Normal overlap | 21883 | 21251 | 97.11 | 182 | 0.83 | 450 | 2.06 |
| High occlusion | 8352 | 6806 | 81.49 | 601 | 7.2 | 945 | 11.31 |
| Moderate occlusion | 14552 | 13076 | 89.86 | 586 | 4.03 | 890 | 6.11 |
| Normal occlusion | 23523 | 22789 | 96.88 | 296 | 1.26 | 438 | 1.86 |

To intuitively show the effect of illumination variation on the performance of TC-YOLO, we collected two images of the same tea chrysanthemum from different angles for testing. The image on the left was collected under weak light, while that on the right was collected under Normal light. The results are shown in Fig. 8.

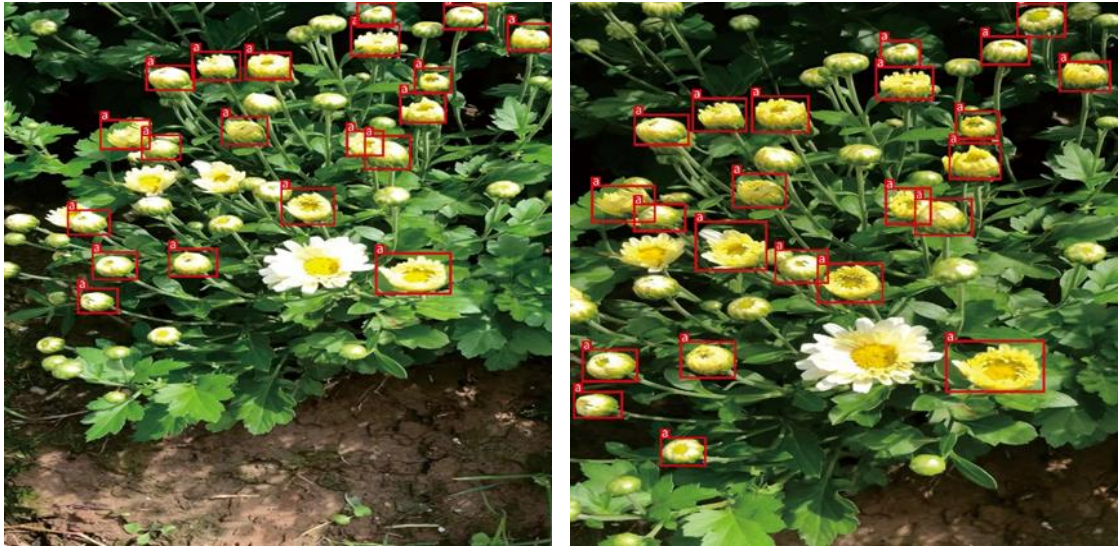

**Fig. 8.** The detection performance of the same tea chrysanthemum under different light intensities. The undetected red box on the left image is manually marked. It can be clearly seen that light intensity plays a major role on the performance of TC-YOLO.

*3.4. Comparisons with State-of-the-Art Detection Methods*

To comprehensively verify the performance of the proposed method, We have compared the proposed TC-YOLO with the latest object detection technologies based on feature fusion, including FPN-based RetinaNet models; EfficientDet (Cao et al., 2020) models, based on EfficientB + BiFPN; multi-level feature pyramid network (MLFPN)-based M2Det (Zhang&Li, 2020) models; YOLOv3 models based on DarkNet53 + FPN; modified sine–cosine algorithm (MSCA)-based VGG16 + PFPNet (Wang et al., 2019) models; RefineDet (Zhang et al., 2018) models, based on Advanced RISC Machine (ARM) + object detection module (ODM); YOLOv4 models based on CSPDarknet53; and YOLOv5 models based on CSPDenseNet.

Table 3

Comparisons with state-of-the-art detection methods.

| Method | Backbone | Size | FPS | AP |
|---|---|---|---|---|
| RetinaNet | ResNet101 | 800 × 800 | 4.53 | 71.51 |

| Model | Backbone | Input Size | FPS | AP |
|---|---|---|---|---|
| RetinaNet | ResNet50 | 800 × 800 | 5.33 | 78.66 |
| RetinaNet | ResNet101 | 500 × 500 | 7.26 | 83.82 |
| RetinaNet | ResNet50 | 500 × 500 | 7.88 | 83.26 |
| EfficientDetD6 | EfficientB6 | 1280 × 1280 | 5.23 | 85.67 |
| EfficientDetD5 | EfficientB5 | 1280 × 1280 | 6.22 | 85.54 |
| EfficientDetD4 | EfficientB4 | 1024 × 1024 | 8.02 | 85.06 |
| EfficientDetD3 | EfficientB3 | 896 × 896 | 9.16 | 86.21 |
| EfficientDetD2 | EfficientB2 | 768 × 768 | 11.63 | 86.45 |
| EfficientDetD1 | EfficientB1 | 640 × 640 | 15.29 | 82.09 |
| EfficientDetD0 | EfficientB0 | 512 × 512 | 37.63 | 80.53 |
| M2Det | VGG16 | 800 × 800 | 7.03 | 80.67 |
| M2Det | ResNet101 | 320 × 320 | 16.88 | 75.88 |
| M2Det | VGG16 | 512 × 512 | 21.28 | 73.29 |
| M2Det | VGG16 | 300 × 300 | 42.56 | 70.22 |
| YOLOv3 | DarkNet53 | 608 × 608 | 12.21 | 87.44 |
| YOLOv3(SPP) | DarkNet53 | 608 × 608 | 15.67 | 84.62 |
| YOLOv3 | DarkNet53 | 416 × 416 | 43.23 | 81.23 |
| YOLOv3 | DarkNet53 | 320 × 320 | 46.92 | 75.52 |
| PFPNet (R) | VGG16 | 512 × 512 | 24.43 | 74.98 |
| PFPNet (R) | VGG16 | 320 × 320 | 33.42 | 80.33 |
| PFPNet (s) | VGG16 | 300 × 300 | 41.33 | 81.58 |
| RFBNetE | VGG16 | 512 × 512 | 21.46 | 78.83 |
| RFBNet | VGG16 | 512 × 512 | 36.11 | 78.36 |
| RFBNet | VGG16 | 512 × 512 | 45.49 | 83.58 |
| RefineDet | VGG16 | 512 × 512 | 31.24 | 78.66 |
| RefineDet | VGG16 | 448 × 448 | 43.09 | 76.85 |
| YOLOv4 | CSPDarknet53 | 608 × 608 | 19.28 | 87.62 |
| YOLOv4 | CSPDarknet53 | 512 × 512 | 24.69 | 87.43 |
| YOLOv4 | CSPDarknet53 | 300 × 300 | 46.61 | 84.92 |
| YOLOv5s | CSPDenseNet | 416 × 416 | 47.62 | 89.86 |
| YOLOv5l | CSPDenseNet | 416 × 416 | 42.22 | 88.96 |
| YOLOv5m | CSPDenseNet | 416 × 416 | 36.89 | 87.23 |
| YOLOv5x | CSPDenseNet | 416 × 416 | 32.23 | 85.26 |
| Ours | CSPD+R | 416 × 416 | 47.23 | 92.49 |

First, the experimental results in Table 3 were analyzed from the perspective of AP. The proposed method achieved the highest AP of 92.49%, which is 2.63% higher than the best-performing feature fusion-based region detector YOLOv5s. There may be two reasons for this. One is that the proposed method based on some fusion components and CSPDenseNet+CSPResNeXt alternate fusion can better guide the gradient flow of tea chrysanthemum features, and the other reason is that in the complex unstructured chrysanthemum dataset, adding RFP structure can increase the gradient backflow in model training, which may activate more detailed features in the chrysanthemum image. It is worth mentioning that the detection speed of the proposed method is not the fastest; based on the 416x416 input scale, YOLOv5s (the fastest) is 0.39 FPS quicker than our approach. This may be because the addition of the gradient backflow mechanism slightly increases the reasoning scale of the model. However, compared with the significant improvement of AP, it is totally acceptable to lose some reasoning speed. The qualitative results of TC-YOLO on the chrysanthemum dataset are shown in Fig. 9.

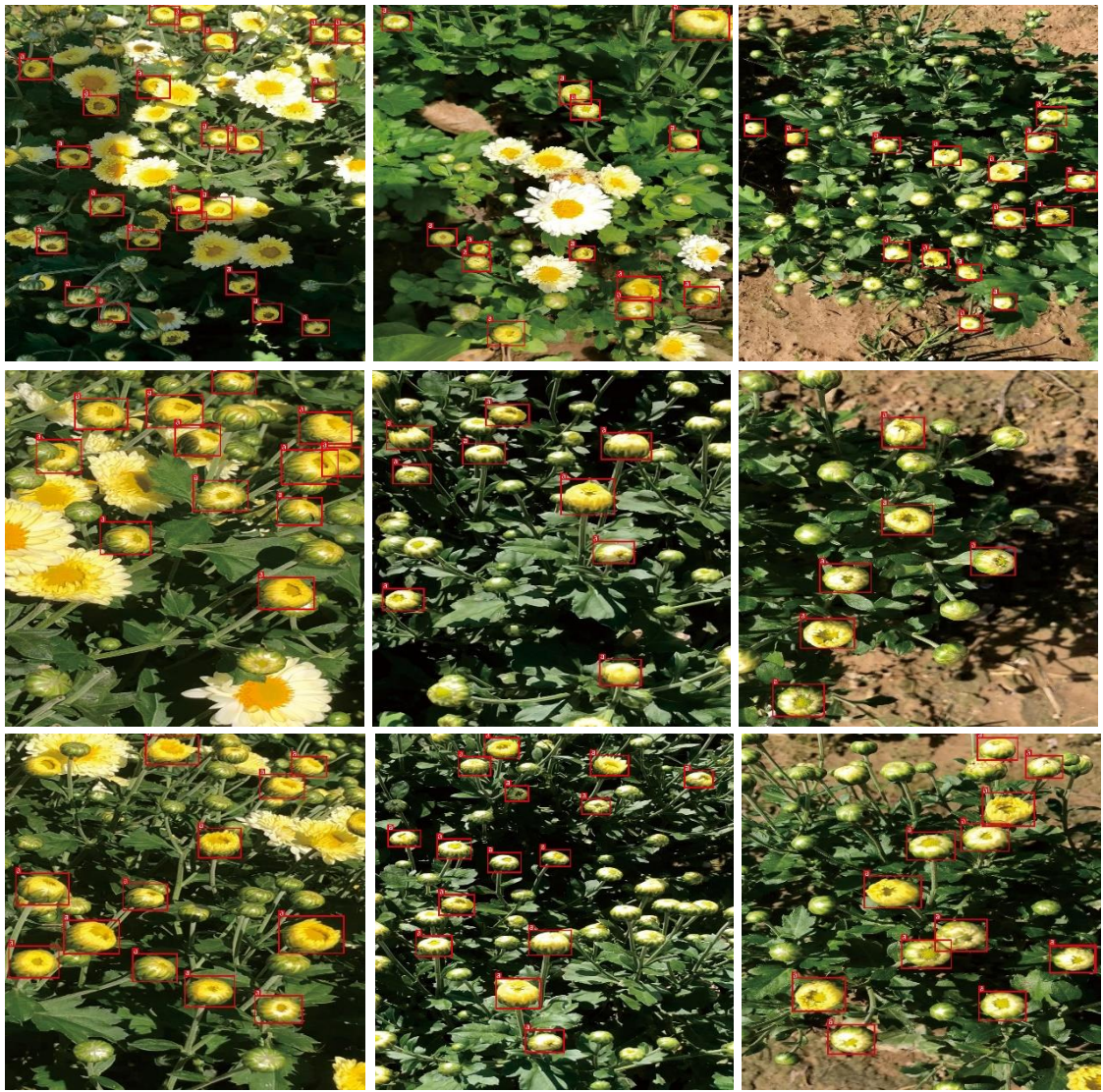

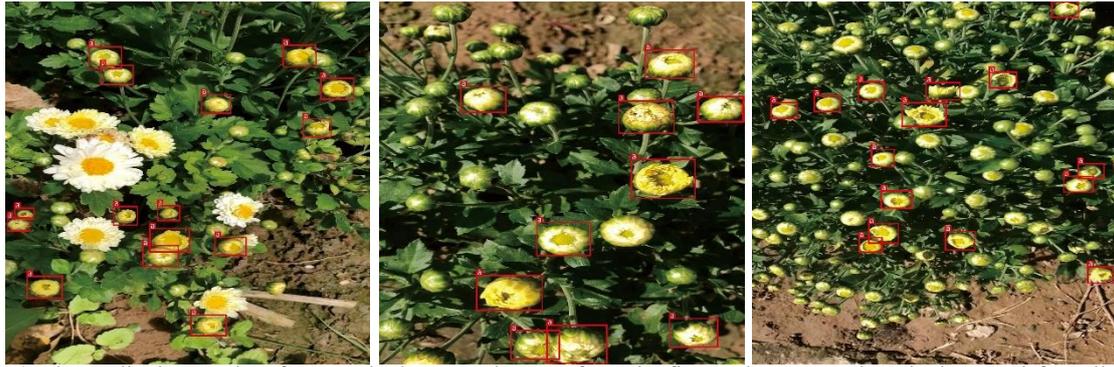

**Fig. 9.** Qualitative results of our method. As can be seen from the figure, the proposed method can satisfactorily detect chrysanthemums in the early flowering stage in complex unstructured scenarios including illumination variation, occlusion and overlap in real time.

## 4. Discussion

In this paper, a lightweight object detection model called tea chrysanthemum - YOLO (TC-YOLO) is proposed, which can adapt to illumination variation, occlusion and overlapping scenarios. In order to realize the detection task, a separate fusion CNN was constructed to fuse different function modules, the purpose of which is to transform chrysanthemum features into common subspace, in which features can be combined linearly or nonlinearly. Moreover, data augmentation methods were applied, and customized specific loss function was used to train these modules. In this way, these function modules can better understand the feature of chrysanthemums at different scales, thus improving the end-to-end performance of the lightweight network in complex unstructured environment. Local transition layer CSPDenseNet was added into different function modules as the main network. The main purpose of the design is to enrich the architecture with gradient combinations and to reduce the amount of calculation. By dividing the feature map of the base layer into two parts and combining them through the proposed cross-stage hierarchy, gradient flow can travel through different network paths. Through the transformation of series and step size, a large correlation difference is generated in the gradient information. Also, the features learned from TC-YOLO are visualized. Fig. 10 shows the network activation of the four input images in the third, tenth, and twentieth convolutional layers. The red, yellow, and blue represent the activated part, the potentially activated part, and the non-activated part respectively. Generally, the regions are different in terms of the degree of activation at different layers for the feature map. Taking the first row in Fig. 10 for instance, starting from the left, the first image is a preprocessed input image. The second image shows the activation of the input image in the third convolutional layer. The upper left region of the feature map has been activated to a high degree, but the red areas are sparse, indicating a low degree of activation. The third image shows the activation of the input image in the tenth convolutional layer. The convergence of red areas is more significant than that in the second image, suggesting that the input image has been activated deeply at the tenth convolutional layer. The fourth image shows the activation of the input image in the twentieth convolutional layer. Obviously, peripheral areas that have not been activated are noticed by the feature fusion mechanism, and features of a large number of peripheral areas are activated.

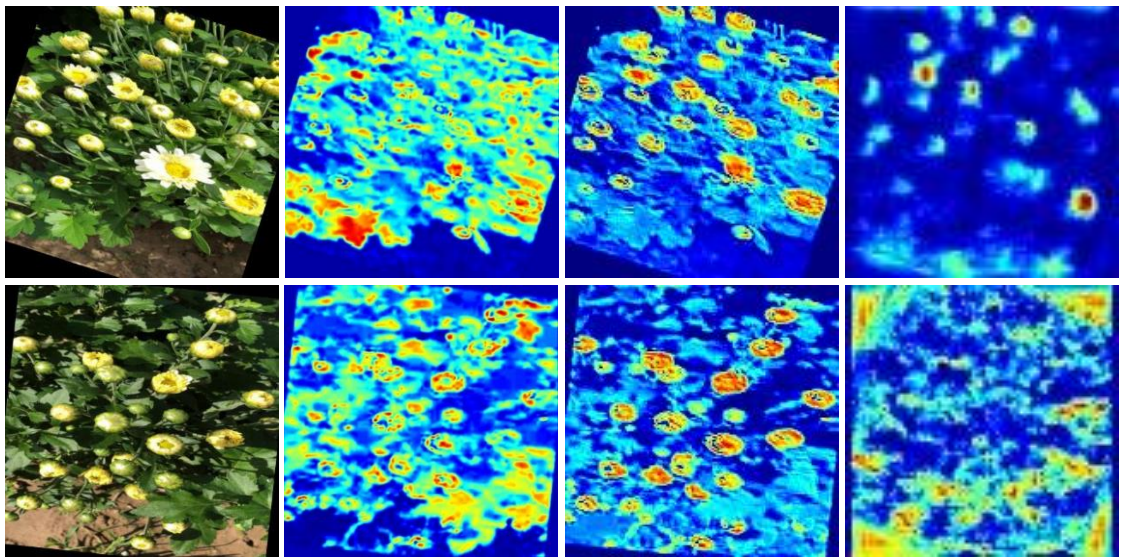

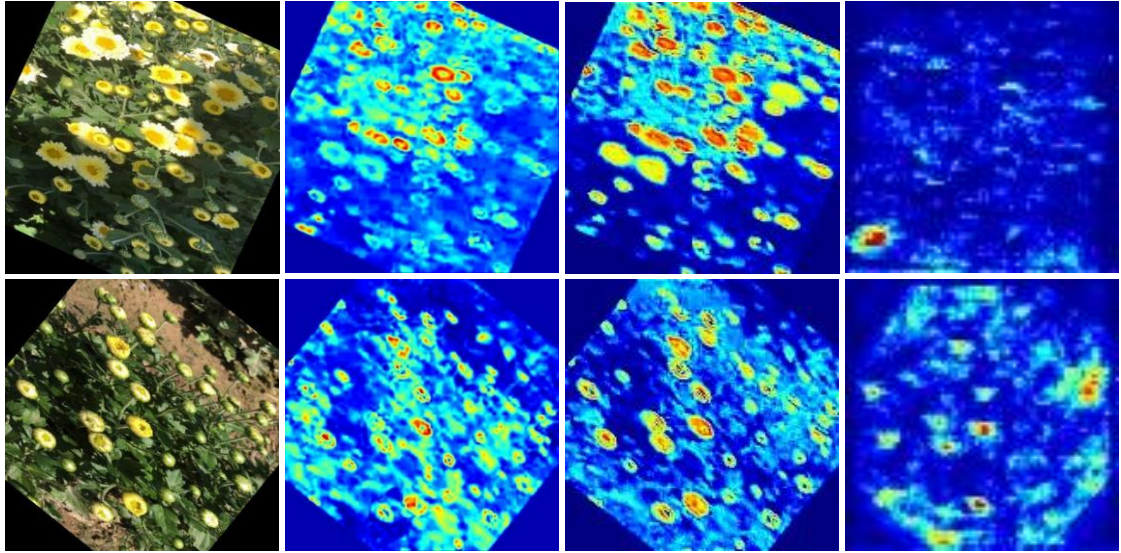

**Fig. 10.** Visualization results of some input images. The first, second, third and fourth columns are for the preprocessed input image, the input image activated at the third convolutional layer, the input image activated at the tenth convolutional layer, and the input image activated at the twentieth convolutional layer, respectively.

Research on machine learning-based chrysanthemum detection is currently scarce worldwide. Traditional machine learning-based chrysanthemum detection has non-negligible drawbacks in terms of inference speed (Kondo et al., 1996; Tarry et al., 2014; Tete&Kamlu, 2017; Warren, 2000; Yang et al., 2018; Yang et al., 2019; Yuan et al., 2018). Among them, the fastest inference speed is the robotic chrysanthemum picking system for Hangzhou white chrysanthemums developed by Yang et al. in 2018, recognizing an image with an inference speed of 0.4 s. Our TC-YOLO detection of an image can reach an inference speed of approximately 21.17 ms (47.23 FPS), which is about 19 times faster than the method of Yang et al. The inference speed of chrysanthemum detection based on deep convolutional networks has improved significantly (Liu et al., 2020; Liu et al., 2019; Van Nam, 2020). The inference speed of the VGG16 and ResNet50-based chrysanthemum detection model proposed by Liu et al. can reach an impressive 10ms, about twice as fast as TC-YOLO. However, the accuracy of this method is only 78%, which is much lower than the detection accuracy of TC-YOLO at 92.49%, probably because VGG16 and ResNet50 extract and use feature information from a single layer for detection, while TC-YOLO not only improves the feature extraction method, but also incorporates a more advanced feature fusion mechanism. It is worth mentioning that most of the current research results for chrysalis detection have been tested in relatively ideal environments, while TC-YOLO was tested in nine unstructured environments. Combining the two factors of inference speed and detection accuracy, there is no doubt about the superiority of TC-YOLO.

The TC-YOLO proposed in this paper has obvious advantages, but also has some shortcomings that need to be addressed. First, from the perspective of performance, the detection performance of TC-YOLO in complex unstructured environment needs further improvement, especially under illumination variation. The challenge of illumination variation for tea chrysanthemum detection is reflected in many aspects. Firstly, illumination variation will change the intuitive feeling of the visual system, of which the most immediate manifestation is the color change. In addition, the change of illumination will make the chrysanthemum look like a mask, which significantly affects the detection accuracy. Currently, there is no public available tea chrysanthemum dataset. Tea chrysanthemum usually matures once a year and typically needs to be picked in the early flowering stage. However, the early flowering period of tea chrysanthemum is very short, which leads to the difficulty to obtain a large amount of data for training networks. Generative Adversarial Networks (GANs) (Espejo-Garcia et al., 2021) are possible to be employed to expand the dataset for training the state-of-the-art deep learning models. These methods are capable of generating realistic images that could increase and diversify the original training datasets. This results in an improved generalization capability of the learned visual classifiers. TC-YOLO is a lightweight (13.6M) fusion network, but whether it can maintain satisfactory speed and accuracy in a mobile embedded device is still not fully investigated. In our future work, we will explore the possibility of GANs to generate additional tea chrysanthemum datasets. Moreover, the pretrained TC-YOLO will be deployed in an embedded platform such as NVIDIA Jetson and its performance will be evaluated to develop the tea chrysanthemum harvesting robot.

5. Conclusions

This paper presents a lightweight CNN architecture called TC-YOLO to detect tea chrysanthemums in the early flowering stage under complex and unstructured environments. We collected 1000 original images (1080 × 1920) as the original dataset. Under the NVIDIA Tesla P100 GPU environment, the AP reached 92.49% in the test dataset, and the inference speed was 47 FPS. The optimal dataset size is 3,000 images for training TC-YOLO, and the best data augmentation strategy is the combination of flip and rotation. We also learn that overlapping has the least impact on the detection of tea chrysanthemums, while lighting has the greatest impact. The proposed method not only achieves the highest AP (92.49%), which is 2.63% higher than the optimal region detector YOLOv5s based on feature fusion, but also keeps a high inference speed (47.23FPS). The proposed lightweight model TC-YOLO has the potential to be integrated into a selective harvesting robot for automatic early flowering tea chrysanthemums harvesting in the future.

CRediT authorship contribution statement

**Chao Qi:** Conceptualization, Methodology, Software, Writing - original draft, Writing - review & editing. **Junfeng Gao:** Conceptualization, Writing - review & editing. **Simon Pearson:** Writing - review & editing. **Helen Harman:** Writing - review & editing. **Kunjie Chen:** Supervision, Writing - review & editing. **Lei Shu:** Supervision, Project administration, Funding acquisition, Writing - review & editing.

**Declaration of Competing Interest**

The authors declare that they do not have any financial or non-financial conflict of interests.

**Acknowledgments**

This work was supported by Lincoln Agri-Robotics as part of the Expanding Excellence in England (E3) Programme.